%% file: main.tex
\documentclass[10pt]{article} % For LaTeX2e
\usepackage[preprint]{tmlr}

\input{math_commands.tex}

\usepackage{hyperref}
\usepackage{url}

\usepackage{times}
\usepackage{latexsym}

\usepackage[T1]{fontenc}
\usepackage[utf8]{inputenc}
\usepackage{wrapfig}

\usepackage{microtype}

\usepackage{inconsolata}

\usepackage{graphicx}

\usepackage{microtype}
\usepackage{graphicx}
\usepackage{subcaption} 

\usepackage{booktabs} % for professional tables
\usepackage{algorithm}
\usepackage{algpseudocode}
\usepackage{algorithmicx}

\usepackage{booktabs} % For better table lines
\usepackage{geometry} % To adjust page margins

\usepackage{amsmath}
\usepackage{amssymb, amsfonts}
\usepackage{multirow}
\usepackage{multicol}
\usepackage{tikz}
\newcommand*\circled[1]{\tikz[baseline=(char.base)]{
            \node[shape=circle,draw,inner sep=1pt] (char) {#1};}}

\makeatletter
\renewcommand{\Function}[2]{%
  \csname ALG@cmd@\ALG@L @Function\endcsname{#1}{#2}%
  \def\jayden@currentfunction{#1}%
}
\newcommand{\funclabel}[1]{%
  \@bsphack
  \protected@write\@auxout{}{%
    \string\newlabel{#1}{{\jayden@currentfunction}{\thepage}}%
  }%
  \@esphack
}
\makeatother

\usepackage{hyperref}

\title{Fast and Cost-effective Speculative Edge-Cloud Decoding with Early Exits}

\author{\name Yeshwanth Venkatesha \email yeshwanth.venkatesha@yale.edu \\
      \addr Department of Electrical Engineering\\
      Yale University
      \AND
      \name Souvik Kundu \email souvikk.kundu@intel.com  \\
      \addr Intel Labs
      \AND
      \name Priyadarshini Panda \email priya.panda@yale.edu\\
      \addr Department of Electrical Engineering\\
      Yale University}

\def\month{MM}  % Insert correct month for camera-ready version
\def\year{YYYY} % Insert correct year for camera-ready version
\def\openreview{\url{https://openreview.net/forum?id=XXXX}} % Insert correct link to OpenReview for camera-ready version

\begin{document}

\maketitle

\begin{abstract}
Large Language Models (LLMs) enable various applications on edge devices such as smartphones, wearables, and embodied robots. However, their deployment often depends on expensive cloud-based APIs, creating high operational costs, which limit access for smaller organizations and raise sustainability concerns.
Certain LLMs can be deployed on-device, offering a cost-effective solution with reduced latency and improved privacy. Yet, limited computing resources constrain the size and accuracy of models that can be deployed, necessitating a collaborative design between edge and cloud.
We propose a fast and cost-effective speculative edge-cloud decoding framework with a large target model on the server and a small draft model on the device.
By introducing early exits in the target model, tokens are generated mid-verification, allowing the client to preemptively draft subsequent tokens before final verification, thus utilizing idle time and enhancing parallelism between edge and cloud.
Using an NVIDIA Jetson Nano (client) and an A100 GPU (server) with Vicuna-68M (draft) and Llama2-7B (target) models, our method achieves up to a 35\% reduction in latency compared to cloud-based autoregressive decoding, with an additional 11\% improvement from preemptive drafting. To demonstrate real-world applicability, we deploy our method on the Unitree Go2 quadruped robot using Vision-Language Model (VLM) based control, achieving a 21\% speedup over traditional cloud-based autoregressive decoding. These results demonstrate the potential of our framework for real-time LLM and VLM applications on resource-constrained edge devices.
\end{abstract}

\section{Introduction}
% Make cost as the main config.
Large Language Models (LLMs) have become pivotal in advancing artificial intelligence, transforming natural language processing (NLP), and enabling a wide range of applications such as chatbots, virtual assistants, robotics, translation, coding, and content generation \cite{zeng2023large, huang2024survey, sun2024survey, zhang2023survey}. Their importance lies in their ability to understand and generate human-like text, making interactions between humans and machines seamless and suggesting potential emergent capabilities \cite{wei2022emergent}. Recent advances include large-scale models like OpenAI’s GPT \cite{radford2019language, brown2020language, achiam2023gpt}, Meta’s LLaMA \cite{touvron2023llama, dubey2024llama}, and Google’s Gemma \cite{team2024gemma}, driving breakthroughs in applications ranging from personalized assistants to complex problem-solving across various domains. 

However, running large models is costly, requiring extensive computational resources for training, often spanning thousands of GPU hours, while inference at scale demands specialized hardware to maintain responsiveness \cite{samsi2023words}.
This creates significant barriers for smaller organizations and researchers who rely on expensive cloud-based APIs; for example, GPT-4.1 text generation costs \$2.00/1M input tokens and \$8.00/1M output tokens at the time of writing this paper.\footnote{OpenAI's API pricing as of May 2025.}

A potential solution is deploying LLMs on edge devices, which offers benefits like low latency, faster customization, and enhanced privacy in addition to cost-effectiveness. This is especially critical for real-time robotics applications, where decisions must be made on the fly, and the server cost can add up. For example, robotic platforms such as the Unitree Go2 quadruped are being equipped with language interfaces for real-world tasks like navigation, object interaction, and instruction following \cite{cheng2024navila}. However, such robots typically run on compute-constrained devices, making it infeasible to host large LLMs locally. For instance, the Unitree Go2 is powered by a Jetson Orin board with 16GB of unified memory, which is insufficient to run models over 10B parameters that require over 40GB of memory.
Efficient decoding strategies like speculative decoding provide cost-effective solutions to bridge this gap. Speculative decoding \cite{leviathan2023fast} uses a smaller model to generate tokens quickly, which are then verified by a larger model in parallel, significantly speeding up LLM inference. Despite its success on standalone machines, the application of speculative decoding on edge devices remains underexplored.
% There is a cost advantage of performing speculative decoding as compared to generating all tokens from a larger model.
% MEI (Mobile Edge Intelligence) provides a solution with AI capabilities at the network edge (6G base stations) that works as a middle ground between LLM on-device and LLM based on the cloud \cite{qu2024mobile}.
% While this is a challenge today given the capacity of the current network, several works have shown the vision of having edge LLMs on 6g networks with projected network speed up to 10 Tbps \cite{banafaa20236g, lin2023pushing, xu2024large, friha2024llm, qu2024mobile, zhang2024edgeshard}. 

% \begin{figure}[h]
%     \centering
%     \includegraphics[width=0.9\linewidth]{figures/motivation_fig_sketch.png}
%     \caption{Motivation }
%     \label{fig:motivation}
% \end{figure}

\begin{figure*}[t]
\vspace{-8mm}
    \centering
    \includegraphics[width=\linewidth]{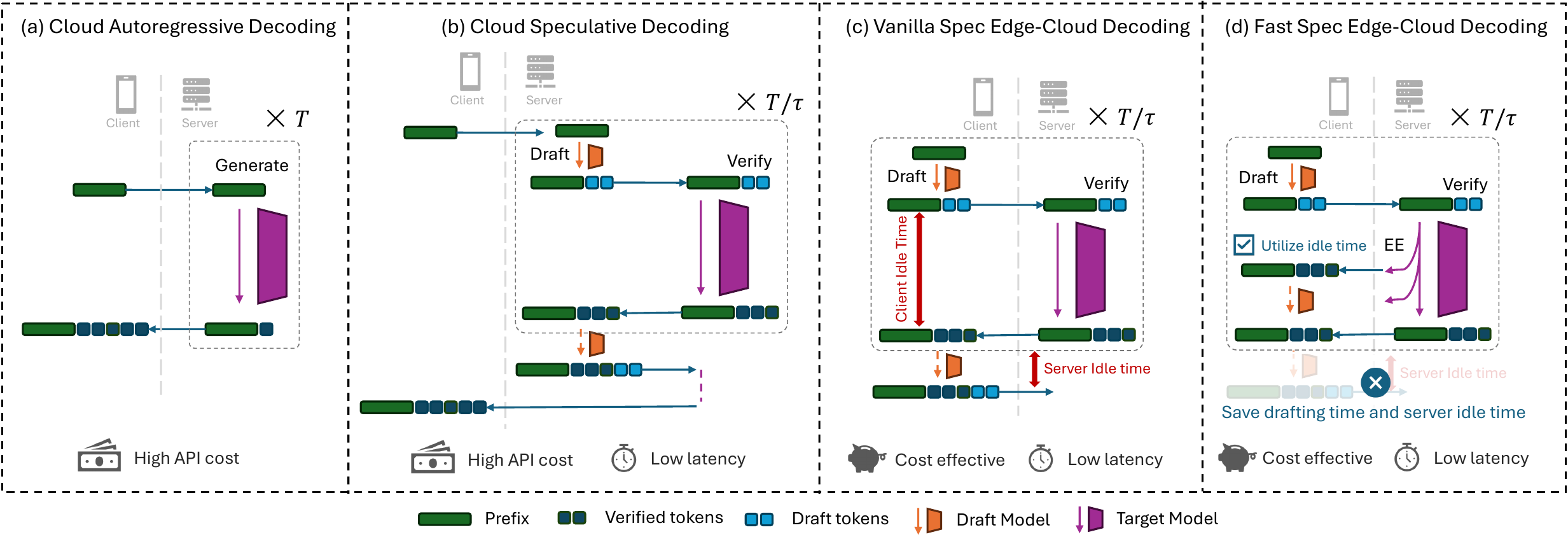}
    % \vspace{-2mm}
    
    \caption{Illustration of traditional cloud-based autoregressive decoding versus cloud-based speculative decoding, vanilla speculative edge-cloud decoding, and the proposed preemptive drafting mechanism.}
    \label{fig:motivation}
\end{figure*}
% \begin{figure}[h!]
%     \centering
%     \includegraphics[width=0.8\linewidth]{figures/motivation_result_v1.pdf}
%     \caption{Projected speedup from autoregressive cloud-based decoding to vanilla speculative edge-cloud decoding, and further speedup expected from our method with early exits in the target model for pre-drafting.}
%     \label{fig:motivation_main_result}
%     \vspace{-3mm}
% \end{figure}

\begin{table*}[t]
    \centering
    \caption{Cost comparison between cloud autoregressive (AR) decoding and cloud speculative decoding (SD) and speculative edge-cloud decoding across different API providers on a set of candidate models. The measurement is based on 1 million requests, each consisting of 100 input tokens and 500 output tokens, assuming a draft length of $\gamma = 4$ tokens and an average of $\tau = 2.5$ accepted tokens per draft.}
    \label{tab:cost_comparison}
    \resizebox{\linewidth}{!}{
    \begin{tabular}{llcccccc}
        \toprule
        \multirow{2}{*}{\textbf{API Provider}} & \multirow{2}{*}{\textbf{Draft/Target}} & \multicolumn{2}{c}{\textbf{Cost (In/Out per 1M tokens)}} & \multicolumn{3}{c}{\textbf{API Cost}} \\
        \cmidrule(lr){3-4} \cmidrule(lr){5-7}
        & & \textbf{Draft} & \textbf{Target} & \textbf{Cloud AR} & \textbf{Cloud SD} & \textbf{Edge-cloud SD} \\
        \midrule
        Together AI & Qwen1.5-0.5B/Qwen1.5-72B & \$0.1/\$0.1 & \$0.9/\$0.9 & \$540 & \$360 (33\% $\downarrow$) & \$270 (\textbf{50\% $\downarrow$})\\
        OpenRouter & llama-3.1-8b/llama-3.1-405b & \$0.02/\$0.05 & \$0.9/\$0.9 & \$540 & \$312 (42\% $\downarrow$) & \$270 (\textbf{50\% $\downarrow$}) \\
        Groq & llama3-8b-8192/llama3-70b-8192 & \$0.05/\$0.08 & \$0.59/\$0.79 & \$454 & \$286 (37\% $\downarrow$) & \$217 (\textbf{52\% $\downarrow$}) \\
        OpenRouter & Qwen-2-VL-7B/Qwen-2-VL-72B & \$0.2/\$0.2 & \$0.7/\$0.7 & \$420 & \$390 (7\% $\downarrow$) & \$210 (\textbf{50\% $\downarrow$}) \\
        \bottomrule
    \end{tabular}

    }
    \vspace{-4mm}
\end{table*}

% Cloud AR: 540.0
% Cloud SD: 360.0
% EC SD: 270.0
% SD Saving: 33.333333333333336%
% EC Saving: 50.0%

% Cloud AR: 540.0
% Cloud SD: 312.0
% EC SD: 270.0
% SD Saving: 42.22222222222222%
% EC Saving: 50.0%

% Cloud AR: 454.0
% Cloud SD: 286.0
% EC SD: 217.0
% SD Saving: 37.00440528634361%
% EC Saving: 52.202643171806166%

% Cloud AR: 420.0
% Cloud SD: 390.0
% EC SD: 210.0
% SD Saving: 7.142857142857143%
% EC Saving: 50.0%

In this work, we propose a novel \textbf{speculative edge-cloud decoding method} to enable fast and cost-effective LLM inference at the edge. As shown in Fig. \ref{fig:motivation}(a), traditional cloud-based autoregressive decoding takes a prompt from the client and performs $T$ forward passes on the target model to generate $T$ tokens, incurring an API cost proportional to $T$. Speculative decoding on the cloud (Fig. \ref{fig:motivation}(b)) reduces target model calls by a factor of $\tau$, the number of tokens generated per draft-verify round. However, it introduces additional draft model calls which comes with a non-negligible cost \cite{yan2024decoding}. Shifting drafting to the edge can eliminate this cost. Table \ref{tab:cost_comparison} shows potential savings for example model pairs from various API providers.\footnote{API cost as of May 2025 based on https://www.helicone.ai/llm-cost.} Speculative edge-cloud decoding can reduce costs by up to 52\% over cloud autoregressive decoding.

A straightforward edge speculation and cloud verification approach (Fig. \ref{fig:motivation}(c)) suffers from inefficiencies: the client remains idle during server verification, and the server is unutilized while the client drafts tokens. To address this, we propose a \textit{preemptive} drafting mechanism to maximize client-server utilization. As shown in Fig. \ref{fig:motivation}(d), we introduce early exits in the target model to produce verified tokens before full verification. These early tokens enable the client to draft the next set preemptively, a process we call \textit{pre-drafting}. If the final verification confirms the early tokens, the next set of draft tokens is readily available for verification, minimizing idle time and keeping both client and server continuously active.
% This framework is particularly well-suited to robotics, where fast, on-device reasoning must be augmented by high-accuracy cloud verification without incurring latency bottlenecks. Our system allows a robot to use a small local model to generate speculative outputs based on visual or language prompts, and asynchronously validate them using a more powerful model in the cloud. This enables responsive and cost-efficient multimodal interaction in real-world, time-sensitive scenarios such as navigation and manipulation.
% For example, in Fig. \ref{fig:motivation_main_result}, we consider a simple example of generating 8 tokens with Vicuna-68M (draft) and Vicuna-7B (target) models. 
% We plot the latency based on measured average drafting, verification, and communication latencies for NVIDIA Jetson Nano (client) and A100 (server).  Assuming a conservative estimate of generating 8 tokens over 3 draft-and-verify cycles with speculative decoding, we can anticipate a speedup of $\sim$1.5x with vanilla speculative edge-cloud decoding. Our method with early exits can further improve it by $\sim$1.25x.
% In this work, we design a speculative decoding method to enable faster LLM inference at the edge. We house the draft model on-device and the target model on a higher-end machine at the network edge. We perform staged verification of draft tokens in a pipelined fashion to enable faster turnaround time.
Our contributions are summarized as follows:
\begin{itemize}
\setlength{\itemsep}{1pt}
\item We propose a novel framework that splits speculative decoding by hosting the draft model on the edge and the target model on the server, significantly reducing target model API costs.
\item We introduce early exits in the target model to generate verified tokens ahead of full verification, enabling the client to preemptively draft the next tokens, minimizing idle time for both client and server.
\item We conduct a comprehensive evaluation across 6 generation tasks on 3 sets of models. With Vicuna-68M as the draft model and Llama2-7B as the target model, we show an average 35\% latency reduction from autoregressive to vanilla edge cloud speculative decoding and a further speedup of 11\% with our fast decoding method.
\item We demonstrate our approach on a real-world robotics platform (Unitree Go2 equipped with an NVIDIA Jetson Orin), highlighting the applicability of our method for enabling edge-cloud collaborative inference in embodied intelligence applications.
\end{itemize}

\section{Background}
% \subsection{Speculative Decoding}
\textbf{Speculative Decoding}:
Speculative decoding follows a Draft-and-Verify approach, where each step starts with generating multiple candidate tokens, which are then verified by the target LLM in parallel, speeding up inference \cite{leviathan2023fast}.
Formally, given an input prefix \(x_{0:t}\), and a target model \(\mathcal{M}_{q}\), a smaller draft model \(\mathcal{M}_{p}\) generates the next \(\gamma\) tokens $x_{t: t + \gamma}$ and their corresponding probability distribution $p_{t: t + \gamma}$ autoregressively:
\begin{equation}
    x_{t: t + \gamma}, p_{t: t + \gamma} = \textsc{Draft}(\mathcal{M}_{p}, x_{0:t})
    \label{eq:draft}
\end{equation}
The target model \(\mathcal{M}_{q}\) verifies these tokens and decides how many to accept denoted by \(\delta\) (\(\delta \leq \gamma\)), then produces the next token:
\begin{equation}
    x_{t: t + \delta + 1} = \textsc{Verify}(\mathcal{M}_{q}, x_{t:t + \gamma}, p_{t: t + \gamma})
    \label{eq:verify}
\end{equation}
The process repeats with the input prefix extended to \(t + \delta + 1\) and passed back to the draft model for the next round.

% \subsection{Early Exit in Large Language Models}
\noindent
\textbf{Early Exit in Large Language Models}:
Early exit strategies improve the efficiency of LLMs by terminating the generation process early if a sufficiently confident output is identified \cite{panda2016conditional, chen2023ee}.
Given an LLM \(\mathcal{M}\) with \(L\) layers and an input sequence \(x_{1:t}\), the hidden state at each layer \(l\) is computed as:  
\begin{equation}
    h^{(l)} = f^{(l)}(h^{(l-1)}, x_{1:t}),
\end{equation}  
where \(h^{(0)}\) is the input embedding.  
At each layer \(l\), the model calculates logits by passing the hidden state through a language model (LM) head, denoted as 
\(\mathbf{z}^{(l)} = \textsc{LMHead}(h^{(l)})\). It also computes a confidence score \(S^{(l)}\) based on the softmax probability:

\begin{equation}
S^{(l)} = \max \left( \text{softmax}(\mathbf{z}^{(l)}) \right).
\label{eq:confidence}
\end{equation}
The model exits early at layer \(l'\) if the confidence score exceeds a predefined threshold, \(S^{(l')} \geq \tau\), and the next token \(x_{t+1}\) is sampled from \(\text{softmax}(\mathbf{z}^{(l')})\). We leverage this mechanism to generate early verified tokens in the target model, which are used to preemptively produce the next set of draft tokens.

% This early exit mechanism allows the model to save computation by skipping unnecessary layers while still producing accurate predictions. If no layer meets the confidence criterion, the output from the final layer \(L\) is used by default.

\section{Methodology}  
% \subsection{Algorithm Design: Client-Server Speculative Decoding - !RENAME}  
\begin{figure*}[t!]  
\vspace{-8mm}
    \centering  
    \includegraphics[width=0.85\linewidth]{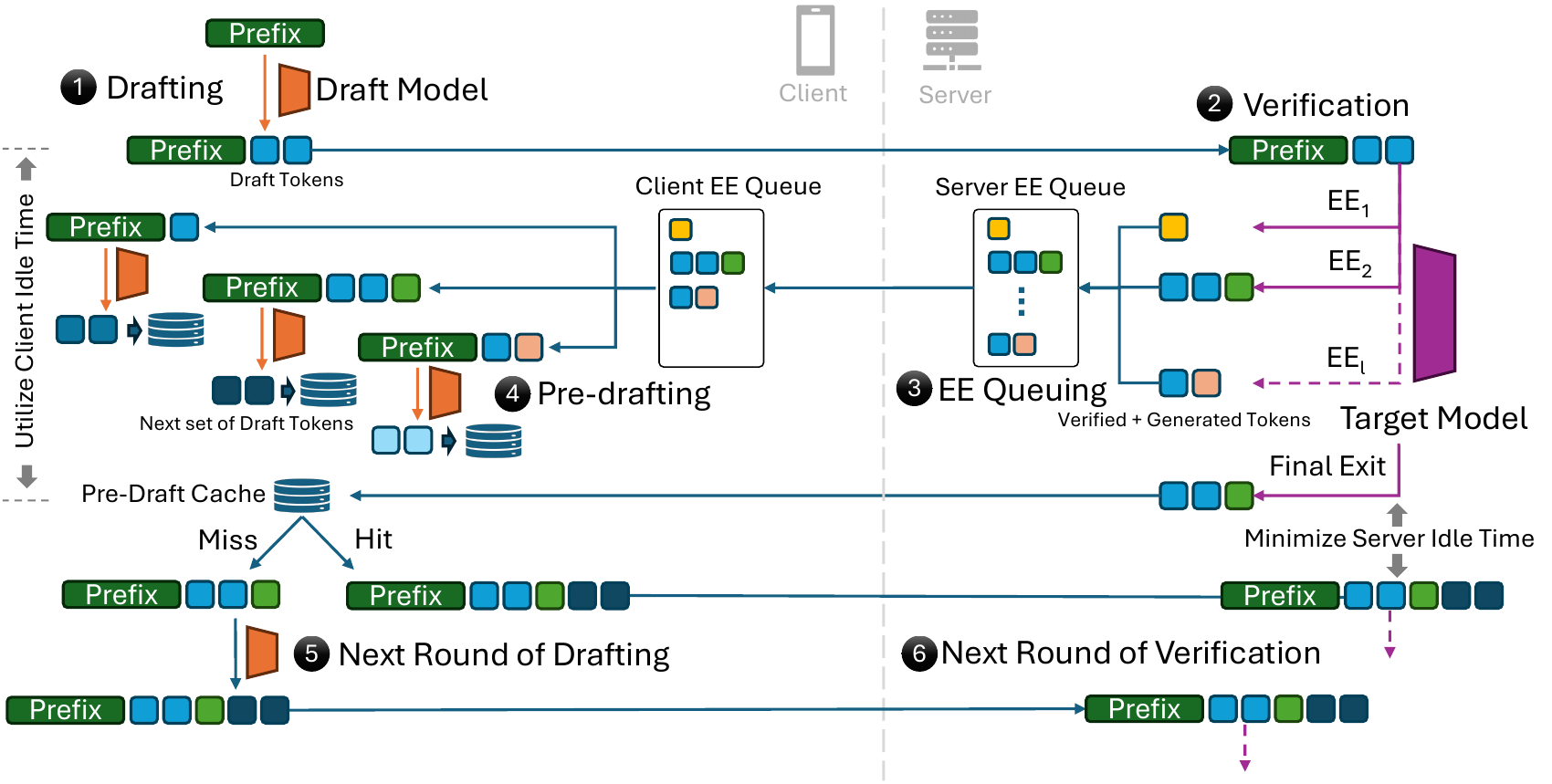}  
    \caption{Illustration of our proposed approach. 
    Given a prefix, the client generates two draft tokens and sends them to the server. The server verifies them using a target model with early exits, returning verified tokens and the next generated token. For each early exit, the client \textit{pre-drafts} the next tokens and stores them in the pre-draft cache. If the final output matches a cache entry, the draft tokens are sent immediately, reducing latency.
    }
    \label{fig:main_fig}
    %\vspace{-3mm}
    
\end{figure*}

In our distributed speculative decoding setup,
the client runs a lightweight \textit{draft model}, denoted as \(\mathcal{M}_{p}\), while the server hosts a large \textit{target model} \(\mathcal{M}_{q}\). As shown in Fig. \ref{fig:main_fig}, the algorithm takes the following sequence of steps:

\textbf{Step \circled{1}:} Given a prefix sequence \(x_{0:t} = \{x_0, x_1, \ldots, x_t\}\), the client uses the draft model \(\mathcal{M}_{p}\) to predict a sequence of \(\gamma\) draft tokens (Eq. \ref{eq:draft}). 
% \begin{equation}  
%     x_{t:t + \gamma} = \{x_{t+1}, x_{t+2}, \ldots, x_{t+\gamma}\} \sim \mathcal{M}_{p}(x_{0:t}).  
% \end{equation}  
These draft tokens $x_{t: t + \gamma}$, along with their probability distributions $p_{t:t + \gamma}$ are transmitted to the server for verification by the target model \(\mathcal{M}_{q}\). 

% Upon receiving the draft tokens, the target model produces a verified sequence:  
% \begin{equation}  
%     x_{t:t + \delta + 1} = \{x_{t+1}, \ldots, x_{t+\delta}, x_{t+\delta+1}\},  
% \end{equation}  
% where \(\delta \leq \gamma\) indicates the number of tokens accepted by the target model, and one additional token \(x_{t+\delta+1}\) is generated to extend the sequence.  

% Multi-Exit Target Model.
\textbf{Step \circled{2}:} The target model \(\mathcal{M}_{q}\) is designed with multiple early exits, denoted as \(\mathcal{M}_{q}^{(1:n)}\). Each early exit \(i \in \{1, \ldots, n\}\) performs a verification step on the draft tokens (Eq. \ref{eq:verify}) and generates the next token. For example, if the early exit $i$ accepts $\delta^{(i)}$ tokens and generates the next token, the total generated tokens would be:  
\begin{equation}  
    x_{t:t + \delta^{(i)} + 1}^{(i)} = \textsc{Verify}(\mathcal{M}_q^{(i)}, x_{t:t + \gamma}, p_{t: t + \gamma})
\end{equation}
Here, $\delta^{(i)}$ denotes the number of draft tokens accepted by early exit $i$.

\textbf{Step \circled{3}:} Given that the communication channel is typically the bottleneck, early exit outputs are queued in the server's early exit queue as soon as they become available and are transmitted to the client sequentially.

\textbf{Step \circled{4}:} The client, in turn, stores the early exit outputs from the server in its own queue and processes each one in a new thread, \textit{preemptively} generating the subsequent set of draft tokens for each early exit. This process is referred to as \textit{pre-drafting}. 
For an early exit \(i\), the newly verified/generated tokens from the server \(x_{t:t + \delta^{(i)} + 1}^{(i)}\) are concatenated with the original prefix \(x_{1:t}\), resulting in a new prefix:  
\begin{equation}
    y^{(i)}_{0:t'} = \text{Concat}(x_{0:t}, x^{(i)}_{t: t + \delta^{(i)} + 1})
\end{equation}
The \textit{pre-draft} tokens represented as \(y^{(i)}_{t':t' + \gamma}\) and their corresponding probabilities \(p^{(i)}_{t':t' + \gamma}\), are then computed as:  
\begin{equation}
    y^{(i)}_{t': t' + \gamma}, p^{(i)}_{t':t' + \gamma} = \textsc{PreDraft}(\mathcal{M}_{p}, y^{(i)}_{0:t'})
\end{equation}
These pre-drafted tokens are subsequently stored in a cache referred to as the pre-draft cache.

\begin{wrapfigure}{r}{0.42\textwidth}
    \centering
    \vspace{-1em}
    \includegraphics[width=0.40\textwidth]{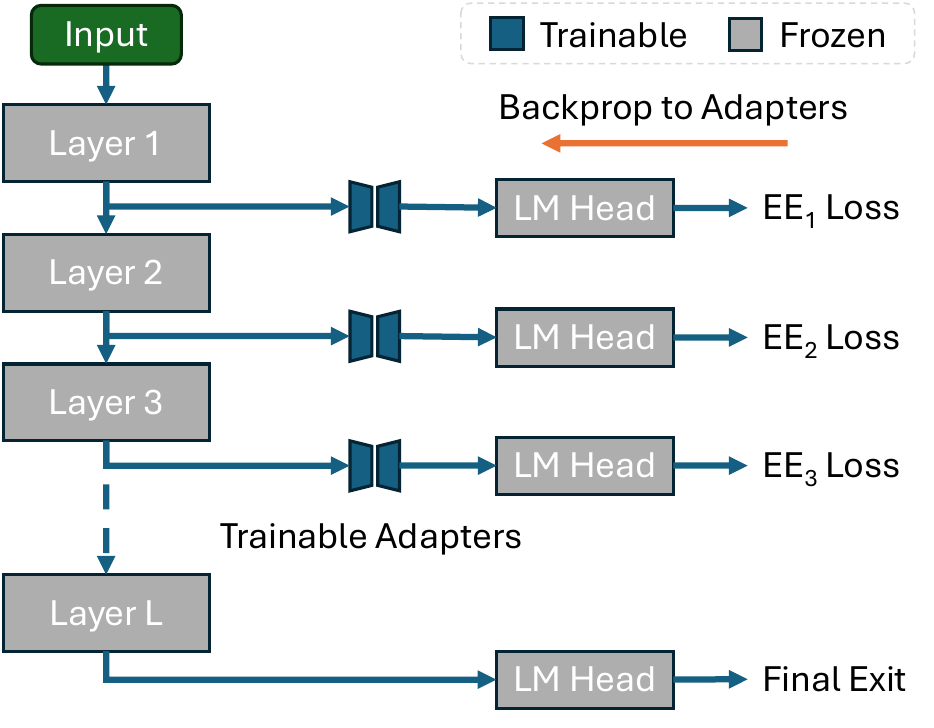}
    \caption{Illustration of training early exit adapters.}
    \label{fig:eeft}
    \vspace{-1.5em}
\end{wrapfigure}

\textbf{Steps \circled{5} \& \circled{6}:} Once the final output \(x_{t:t + \delta + 1}\) from the target model is received, the client checks whether these tokens were already processed in any of the early exits by looking at the pre-draft cache. If there is a hit, the corresponding pre-draft tokens are retrieved from the pre-draft cache and immediately sent to the server for the next round of verification, avoiding any delay. If it is a miss, a new set of draft tokens is generated following the usual drafting process.  The server proceeds with the next round of verification over the new set of draft tokens.

This design enhances efficiency by leveraging the client’s idle time for pre-drafting and reducing the server’s idle time between verification rounds whenever there is a pre-draft cache hit. Importantly, the output is identical to that of standard speculative decoding since all tokens are verified at the final exit of the target model, guaranteeing no loss in accuracy.
For detailed system design and pseudocode please refer to Appendix \ref{section:appendix_system_design}.

\noindent
\textbf{Early Exit Training}:
We add adapter layers after each layer of the target model to train the early exits, as shown in Fig.~\ref{fig:eeft}. Each adapter connects to the language model (LM) head, and its loss is backpropagated to update only that adapter. This minimizes trainable parameters while preserving the original model.

For language generation models, we train early exit adapters on the publicly available ShareGPT conversation dataset (hf:RyokoAI/ShareGPT52K) using a single NVIDIA A100 GPU with 80GB of VRAM. We fine-tune three models—Vicuna-7B, Vicuna-13B, and Llama2-7B—for 10 epochs each, using a batch size of 1 and a learning rate of 1e-4.
Additionally, we train early exit adapters for a vision-language model based on Qwen2VL-7B using the Spacellava dataset (hf:remyxai/vqasynth\_spacellava) which is generated by open source implementation of SpatialVLM \cite{chen2024spatialvlm}.
Table~\ref{tab:eeft} summarizes the number of early exits, total training time, and the number of trainable parameters for each model. Note that the context length was reduced during training to ensure compatibility with the memory limitations of a single A100 GPU.

\begin{table*}[t]
\vspace{-8mm}
    \caption{Early Exit training details. \# Params and \% Params denote the total number of trainable adapter parameters and their fraction compared to total model parameters respectively.}
    \label{tab:eeft}
    \centering
    \resizebox{\linewidth}{!}{%
    \begin{tabular}{lccccc}
    \toprule
      Model   & \# Exits & \#  Params & \% Params & Context & GPU Hours \\
    \midrule

       lmsys/Vicuna-7B-v1.3 \cite{zheng2023judging}  & 31 & 101M & 1.48 & 1600 & 117 \\
       % trainable params: 101,068,800 || all params: 6,839,484,416 || trainable%: 1.4777

        lmsys/Vicuna-13B-v1.3 \cite{zheng2023judging} & 39 & 158M & 1.20 & 800 & 122 \\
       %trainable params: 157,916,160 || all params: 13,173,780,480 || trainable%: 1.1987

       meta-llama/Llama-2-7B-hf \cite{touvron2023llama}  & 31 & 101M & 1.48 & 1600 & 119 \\
       % trainable params: 101,068,800 || all params: 6,839,484,416 || trainable%: 1.4777

       Qwen/Qwen2-VL-7B-Instruct \cite{wang2024qwen2} & 27 & 88M & 1.02 & 1600 & 136 \\

    \bottomrule
    \end{tabular}
    }
    % \vspace{-3mm}
    
\end{table*}

% \begin{table*}[h!]
%     \centering
%     \caption{Notations used in our analysis.}
%     \label{tab:notations}
%     \begin{tabular}{@{}ll@{}}
%         \toprule
%         \textbf{Notation} & \textbf{Description} \\ \midrule
%         $T_q$ & Time required for a single forward pass of the target model $\mathcal{M}_q$ \\
%         $T_p$ & Time taken for a forward pass of the draft model $\mathcal{M}_p$ \\
%         $T_c$ & Communication latency between the client and server \\
%         $c = T_p/T_q$ & Latency ratio between draft and target model\\
%         $\tau$ & Average number of tokens generated per draft-verification round \\
%         $n$ & Number of tokens generated \\
%         $\gamma$ & Draft tokens generated per draft-verification round \\
%         $T_r$ & Latency incurred due to thread synchronization in cases of cache hits \\
%         $r$ & Cache miss rate \\ \bottomrule
%     \end{tabular}
% \end{table*}
% Add cost data

\begin{table}[t]
% \vspace{-5mm}
    \centering
    \caption{Notations used in our analysis.}
    \label{tab:notations}
    \resizebox{\linewidth}{!}{%
    \begin{tabular}{@{}rp{0.79\linewidth}@{}}
        \toprule
        \textbf{Notation} & \textbf{Description} \\ \midrule
        $T_p$ : & Time for a single forward pass of the draft model $\mathcal{M}_p$ \\ 
        
        $T_q$ : & Time for a forward pass of the target model $\mathcal{M}_q$ \\ 
        $T_c$ : & Communication latency between the client and server \\ 
        $c$ : & Latency ratio between the draft and target models ($T_p / T_q$) \\ 
        $\gamma$ : & Number of draft tokens \\ 
        
        $\tau$ : & Effective number of tokens generated per draft-verify round (\# accepted tokens + 1 generated). \\ 
        $n$ : & Total number of tokens generated \\ 
        $r$ : & Cache miss rate \\ 
        $T_r$ : & Latency of thread synchronization on cache hit \\ 
        \bottomrule
    \end{tabular}
    }
    \vspace{-3mm}
    
\end{table}

\section{Experiments}\label{section:analysis}
\textbf{Models and Benchmarks:} Following the standard speculative decoding literature \cite{li2024eagle}, we evaluate our method on three model sets: Vicuna-68M/Vicuna-7B, Vicuna-160M/Vicuna-13B, and Vicuna-68M/Llama2-7B. 
We show the experiments on 6 standard generative task benchmarks spanning conversation \cite{zheng2023judging}, code generation \cite{chen2021evaluating}, mathematical reasoning \cite{cobbe2021training}, instruction following \cite{taori2023stanford}, summarization \cite{nallapati2016abstractive}, and question-answering tasks \cite{kwiatkowski2019natural}.

\noindent
\textbf{Server Side Hardware}:
We utilize a high performance computing cluster node equipped with a single A100 GPU with 80GB VRAM, 16 CPU cores, and 8GB of CPU memory per core as our server.

\noindent
\textbf{Client Side Hardware:}
We demonstrate our system on two types of client devices:
% %\vspace{-3mm}
\begin{enumerate}
    \setlength{\itemsep}{1pt}
    \item NVIDIA Jetson Nano: A compact AI development board tailored for edge computing. It includes a quad-core ARM Cortex-A57 CPU, a 128-core Maxwell GPU, and 4GB of LPDDR4 RAM shared between the CPU and GPU. With a performance of up to 472 GFLOPs, the Jetson Nano is ideal for edge applications.
    \item Cluster Node with RTX 2080 Ti: This setup features a single RTX 2080 Ti GPU with 12GB VRAM, an 8-core CPU, and 4GB of RAM per core, providing a more powerful alternative for our experiments.
\end{enumerate}

% https://www.nvidia.com/en-us/autonomous-machines/embedded-systems/jetson-nano/product-development/
% https://www.nvidia.com/en-us/autonomous-machines/embedded-systems/jetson-orin/

\noindent
\textbf{Communication:} Communication between cluster nodes is facilitated by InfiniBand high-speed interconnect. Ethernet is used for communication between the cluster node and the Jetson Nano.

\begin{wraptable}{r}{0.55\textwidth}
    \centering
    \vspace{-1.3em} % optional: shifts table upward slightly
    \caption{Latency calculation. AR denotes cloud-based autoregressive decoding. SD and FSD refer to vanilla and fast speculative edge-cloud decoding, respectively.}
    \label{tab:latency_calc}
    \begin{tabular}{ll}
        \toprule
        \textbf{Method} & \textbf{Latency} \\
        \midrule
        Cloud AR & $2T_c + nT_q$ \\
        Edge-Cloud SD & $\frac{n}{\tau}(2T_c + \gamma T_p + T_q)$ \\
        Edge-Cloud FSD & $\frac{n}{\tau}(2T_c + r\gamma T_p + (1 - r)T_r + T_q)$ \\
        \bottomrule
    \end{tabular}
    \vspace{-0.8em} % optional: reduce space below the table
\end{wraptable}

\noindent
\textbf{Latency Calculation}:
Table~\ref{tab:notations} summarizes the notations used in our analysis. Key variables include $T_q$, $T_p$, and $T_c$, representing latencies of the target model, draft model, and communication, respectively. Speculative decoding metrics include $\tau$, the effective tokens per draft-verification round. Hyperparameters are $n$, the total tokens, and $\gamma$, tokens per draft-verification round. Additional factors for our fast speculative decoding method include the cache miss rate $r$ and synchronization latency $T_r$ for cache hits.
The latency calculation for autoregressive (AR) decoding, vanilla speculative edge-cloud decoding (SD), and our fast speculative edge-cloud decoding (FSD) are presented in Table~\ref{tab:latency_calc}. The latency of FSD method depends on the cache miss rate, $r$. 
In case of a cache hit, threads must synchronize, incurring a latency of $T_r$. 
Unless otherwise specified, we use $\gamma = 4$ and $n = 200$ in our experiments.

\subsection{System Metrics}
% \begin{figure*}
%     \centering
%     \includegraphics[width=0.7\linewidth]{figures/profiling_sketch.png}
%     \caption{Time spent at each thread and between server and client. Color coded to show time spent in each .}
%     \label{fig:profiling}
% \end{figure*}

% In Table \ref{tab:system_metrics}, we provide the system metrics. In particular we report the drafting latency ($\gamma T_p$) at $\gamma = 4$ that is the latency for generating 4 draft tokens on the client. And verification latency, that is the latency of one forward pass of the target model required for verifying the tokens in the server. Communication latency and the maximum GPU VRAM used. We report the maximum early exit threads that the system can handle. 
% In Table~\ref{tab:system_metrics}, we present the system metrics, which are dataset-agnostic. Specifically, we report the drafting latency (\(\gamma T_p\)) at \(\gamma = 4\), representing the time required to generate 4 draft tokens on the client. We also include the verification latency (\(T_q\)), defined as the time taken for a single forward pass of the target model to verify the draft tokens and generate the next token. Additionally, the communication latency (\(T_c\)) is reported. 
% The efficiency of speculative decoding depends on the draft model being significantly faster than the target model, typically quantified by the latency ratio \(c = T_p / T_q\). We include this ratio as part of the metrics.

We report the system metrics in Table~\ref{tab:system_metrics}, including drafting latency (\(\gamma T_p\)) at \(\gamma = 4\), verification latency (\(T_q\)), the latency ratio \(c\), and communication latency (\(T_c\)).  
On the Jetson Nano, drafting is about three times slower and communication twice as slow as on a cluster node with an RTX GPU. 
We report the maximum GPU VRAM and the number of early-exit threads supported by the system. On the RTX-equipped node, the system handles up to 30 threads for the Vicuna-68M model, but GPU VRAM (12 GB) limits the Vicuna-160M model to 15 threads before encountering an out-of-memory (OOM) error. On the Jetson Nano, both CPU threading and RAM are bottlenecks. The maximum number of threads is capped at 15 for the Vicuna-68M model, while the 4 GB memory limit allows only 7 threads for the Vicuna-160M model.

\begin{table}[h!]
  \centering
  \caption{Average system metrics that are dataset agnostic.}
  \label{tab:system_metrics}
  % \vskip 0.1in
  \resizebox{\linewidth}{!}{%
    \begin{tabular}{l cc cc cc}
    \toprule
    \multirow{2}{*}{\textbf{Metric}} 
    & \multicolumn{2}{c}{\textbf{Vicuna-68m/Vicuna-7B}} 
    & \multicolumn{2}{c}{\textbf{Vicuna-160m/Vicuna-13B}} 
    & \multicolumn{2}{c}{\textbf{Vicuna-68m/Llama2-7B}} \\
     & Jetson & RTX & Jetson & RTX & Jetson & RTX \\
    \midrule
      Drafting Latency ($\gamma T_p, \gamma = 4$) & 334ms & 102ms & 1596ms & 555ms &  301ms & 99ms \\
      Verification Latency ($T_q$) & 497ms & 442ms & 616ms & 618ms & 522ms & 467ms \\
      Latency Ratio ($c = T_p/T_q$) & 0.17 & 0.06 & 0.65 & 0.22 & 0.14 & 0.05 \\
      Communication Latency ($T_{c}$) & 95ms & 42ms & 91ms & 46ms & 96ms & 47ms \\
      Max GPU memory   & 1.7G & 3.2G  & 3.5G & 8.9G & 1.7G & 3.2G \\
      Num EE Threads   & 15 & 30 & 7 & 15 & 15 & 30 \\

    \bottomrule
    \end{tabular}%
  }
    % \vspace{-4mm}
  
\end{table}

\subsection{Speedup Results}

% We might not need AR to SD. Reconsider!!!
\begin{table*}[t]
\vspace{-8mm}
  \centering
  \caption{Speedup evaluation on standard language benchmark datasets.}
  \label{tab:main_speedup}
  % \vskip 0.1in
  \resizebox{\linewidth}{!}{%
    \begin{tabular}{l l cc cc cc}
    \toprule
    \multirow{2}{*}{\textbf{Benchmark}} & \multirow{2}{*}{\textbf{Metric}} 
    & \multicolumn{2}{c}{\textbf{Vicuna-68m/Vicuna-7B}} 
    & \multicolumn{2}{c}{\textbf{Vicuna-160m/Vicuna-13B}} 
    & \multicolumn{2}{c}{\textbf{Vicuna-68m/Llama2-7B}} \\
    & & Jetson & RTX & Jetson & RTX & Jetson & RTX \\
    \midrule
    \multirow{5}{*}{MT-bench} 
      & Speedup AR $\rightarrow$ SD & 0.70x & 1.30x & 0.42x & 0.97x & 1.34x & 2.01x \\
      & Speedup SD $\rightarrow$ FSD & 1.04x & 1.04x & 1.05x & 1.09x & 1.07x & 1.02x \\
      & Avg Tokens $\tau$     & 2.30 & 2.01 & 2.28 & 2.98 & 4.12 & 3.48 \\
      
      & Cache miss rate    & 60.92\% & 20.07\% & 62.49\% & 16.40\% & 36.73\% & 13.38\% \\
      & Avg EE     & 8 & 13 & 9 & 15 & 9 & 11 \\
      
    \midrule
    \multirow{5}{*}{HumanEval} 
      & Speedup AR $\rightarrow$ SD  & 0.79x & 1.42x & 0.47x & 0.85x & 1.53x & 2.09x \\
      & Speedup SD $\rightarrow$ FSD  & 1.04x & 1.03x & 1.06x & 1.15x & 1.22x & 1.07x \\
      & Avg Tokens $\tau$     & 2.04 & 2.66 & 2.14 & 2.04 & 4.16 & 3.83 \\
      
      & Cache miss rate    & 64.73\% & 15.79\% & 57.64\% & 25.16\% & 17.6\% & 1.75\% \\
      & Avg EE     & 7 & 16 & 8 & 14 & 4 & 3 \\
      
    \midrule
    \multirow{5}{*}{GSM8K} 
      & Speedup AR $\rightarrow$ SD  & 0.63x & 1.11x & 0.40x & 0.77x & 1.15x & 1.69x \\
      & Speedup SD $\rightarrow$ FSD  & 1.04x & 1.08x & 1.06x & 1.13x & 1.07x & 1.06x \\
      & Avg Tokens $\tau$     & 2.08 & 1.96  & 2.23 & 2.29 & 3.48 & 3.29 \\
      
      & Cache miss rate    & 61.21\% & 12.40\% & 58.04\% & 18.37\% & 41.87\% & 9.55\% \\
      & Avg EE     & 8 & 13 & 9 & 14 & 7 & 10 \\
      
    \midrule
    \multirow{5}{*}{Alpaca} 
      & Speedup AR $\rightarrow$ SD  & 0.63x & 1.06x & 0.42x & 0.74x & 1.42x & 1.99x \\
      & Speedup SD $\rightarrow$ FSD  & 1.04x & 1.07x & 1.05x & 1.12x & 1.10x & 1.18x \\
      & Avg Tokens $\tau$     & 2.09 & 1.96 & 2.32 & 2.36 & 4.29 & 3.62 \\
      
      & Cache miss rate    & 64.60\% & 19.29\% & 60.94\% & 25.45\% & 36.60\% & 4.28\% \\
      & Avg EE     & 8 & 14 & 8 & 14 & 9 & 2 \\
      
    \midrule
    \multirow{5}{*}{CNN/DM} 
      & Speedup AR $\rightarrow$ SD  & 0.72x & 1.20x & 0.38x & 0.73x & 1.41x & 1.91x \\
      & Speedup SD $\rightarrow$ FSD  & 1.03x & 1.07x & 1.03x & 1.07x & 1.07x & 1.04x \\
      & Avg Tokens $\tau$     & 2.32 & 1.95 & 2.08 & 2.10 & 4.32 & 3.43 \\
      & Cache miss rate  & 70.70\% & 29.40\% & 69.92\% & 38.09\%  & 47.13\% & 13.97\% \\
      & Avg EE     & 10 & 15 & 8 & 16 & 14 & 13 \\
      
    \midrule
    \multirow{5}{*}{NQ} 
      & Speedup AR $\rightarrow$ SD  & 0.65x & 1.10x & 0.46x & 0.82x & 1.26x & 1.93x \\
      & Speedup SD $\rightarrow$ FSD  & 1.02x & 1.06x & 1.04x & 1.12\% & 1.05x & 1.01x \\
      & Avg Tokens $\tau$     & 2.08 & 2.05 & 2.44 & 2.62 & 3.83 & 3.62 \\
      & Cache miss rate    & 71.50\% & 22.59\% & 63.25\% & 30.25\% & 57.11\% & 14.88\% \\
      & Avg EE     & 8 & 14 & 8 & 15 & 13 & 14 \\
      
    \midrule
    \midrule
    \multirow{5}{*}{\textbf{Average}} 
      & \textbf{Speedup AR $\rightarrow$ SD ($\uparrow$)} & 0.69x & 1.20x & 0.42x & 0.94x & 1.35x & \textbf{1.94x} \\
      & \textbf{Speedup SD $\rightarrow$ FSD ($\uparrow$)}  & 1.04x & 1.06x & 1.05x & 1.10x & \textbf{1.11x} & 1.06x \\
      & \textbf{Avg Tokens $\tau$   ($\uparrow$)}  & 2.15 & 2.10 & 2.25 & 2.40 & \textbf{4.03} & 3.54 \\
      
      & \textbf{Cache miss rate ($\downarrow$)}    & 65.61\% & 19.59\% & 62.05\% & 27.95\% & 39.94\% & \textbf{9.63\%} \\
      & \textbf{Avg EE ($\downarrow$)}     & \textbf{8} & 14 & \textbf{8} & 14 & 9 & 9 \\
    \bottomrule
    \end{tabular}%
  }
  \vspace{-5mm}
\end{table*}

% Add table for VLMs

\textbf{Evaluation Metrics:}
Our fast decoding method with early exit is exact, with outputs identical to standard speculative decoding, ensuring \textbf{no loss in accuracy}. 
We define the following metrics to evaluate our method.
% %\vspace{-3mm}
\begin{itemize}
    \setlength{\itemsep}{1pt}
    \item \textbf{Speedup AR $\rightarrow$ SD}: Latency savings of vanilla speculative edge-cloud decoding (SD) compared to cloud based autoregressive (AR) baseline.
    \item \textbf{Speedup SD $\rightarrow$ FSD}: Latency savings of our fast speculative edge-cloud decoding (FSD) compared to the vanilla speculative edge-cloud decoding (SD).
    \item \textbf{Cache miss rate (lower the better)}: Frequency of cache misses, that indicates how often we fail to find the final output in one of the early exits.
    \item \textbf{Average Early Exit (lower the better)}: The average early exit that produces the same output as the final exit.
\end{itemize}
Table~\ref{tab:main_speedup} presents the evaluation metrics on the benchmark datasets. 
%We present the speedup of vanilla speculative edge-cloud decoding over cloud-based autoregressive decoding (Speedup AR~$\rightarrow$SD) and the speedup of our fast speculative decoding over vanilla speculative decoding (Speedup SD$\rightarrow$~FSD).
In addition to the aforementioned evaluation metrics, the effective number of generated tokens per verification ($\tau$) is also reported. Note, $\tau$ remains identical to that of vanilla SD as our FSD method produces identical outputs but it highlights the reduction in API call costs.
% Additionally, we include key metrics specific to our FSD method—Cache miss rate and average early exit.

% \noindent
\textbf{AR $\rightarrow$ SD:}
On average, using RTX, vanilla SD achieves a 1.2x and 1.94x speedup over autoregressive decoding with the Vicuna-68M/Vicuna-7B and Vicuna-68M/Llama2-7B models, respectively. However, it results in a marginal slowdown with Vicuna-160M/Vicuna-13B.
Jetson, being slower at drafting coupled with higher communication cost, reduces the speedup relative to autoregressive decoding, making it slower for Vicuna-68M/Vicuna-7B and Vicuna-160M/Vicuna-13B models, though it achieves a 1.34x speedup with Vicuna-68M/Llama2-7B.
The speedup from autoregressive to SD primarily depends on $c$, $\tau$, and $T_c$, with ideally requiring low values of $c$ and $T_c$ and a high $\tau$. Since Jetson has a high $c$ and $T_c$, it underperforms compared to autoregressive on Vicuna-68M/Vicuna-7B and Vicuna-160M/Vicuna-13B models. In contrast, for Vicuna-68M/Llama2-7B, a lower $c$ combined with a higher $\tau$ yields a 1.35x speedup.

% 301 + 522 + 2*96 for 4.03 tokens
% SD: 1015/4.03 = 251.86 per token, 
% AR: 340
% Ours: 226.9
% 340 to 252 to 227

% In this section, we provide a detailed analysis of the speedup achieved by our decoding method. We break down the different components contributing to the overall latency and provide an understanding on how these interact to affect speedup. 

% \noindent
\textbf{SD $\rightarrow$ FSD:}
Our FSD provides consistent speedup over vanilla SD across all datasets on both RTX and Jetson. On the RTX client, it achieves average speedups of 1.06x, 1.10x, and 1.06x for Vicuna-68M/Vicuna-7B, Vicuna-160M/Vicuna-13B, and Vicuna-68M/Llama2-7B, respectively. Similarly, on Jetson Nano, the speedups are 1.04x, 1.06x, and 1.11x for the same model pairs.
The primary benefit of FSD lies in its \textit{pre-drafting} mechanism, which enables these improvements over vanilla SD. This mechanism’s impact is reflected in the cache miss rate, indicating how often the final output is available through early exits, allowing pre-drafting of the next set of tokens. The average early exit metric further highlights how quickly verified tokens are obtained, enabling efficient generation of subsequent draft tokens.

% \begin{figure}[t]
%     \centering
%     \includegraphics[width=0.7\linewidth]{figures/num_threads_v3.pdf}
%     \caption{Effect of number of early exit threads.}
%     \label{fig:effect_threads}
%     %\vspace{-1mm}
% \end{figure}

\begin{figure}[t]
\vspace{-7mm}
    \centering
    \begin{subfigure}[t]{0.44\linewidth}
        \centering
        \includegraphics[width=\linewidth]{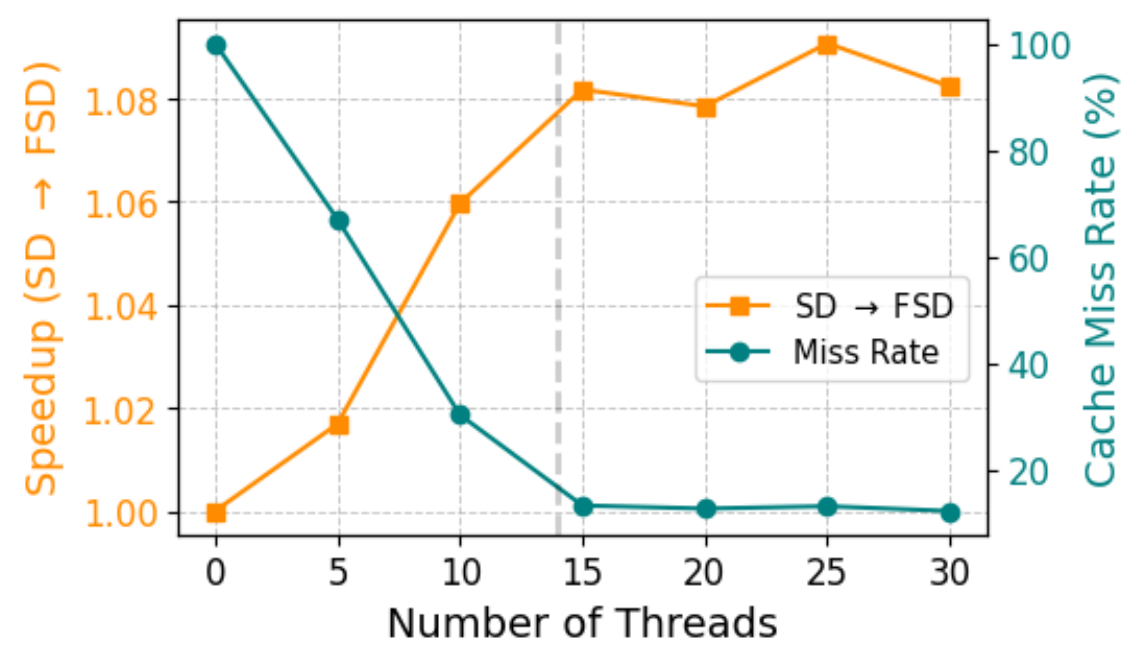}
        \caption{Effect of number of early exit threads.}
        \label{fig:effect_threads}
    \end{subfigure}
    \hfill
    \begin{subfigure}[t]{0.44\linewidth}
        \centering
        \includegraphics[width=\linewidth]{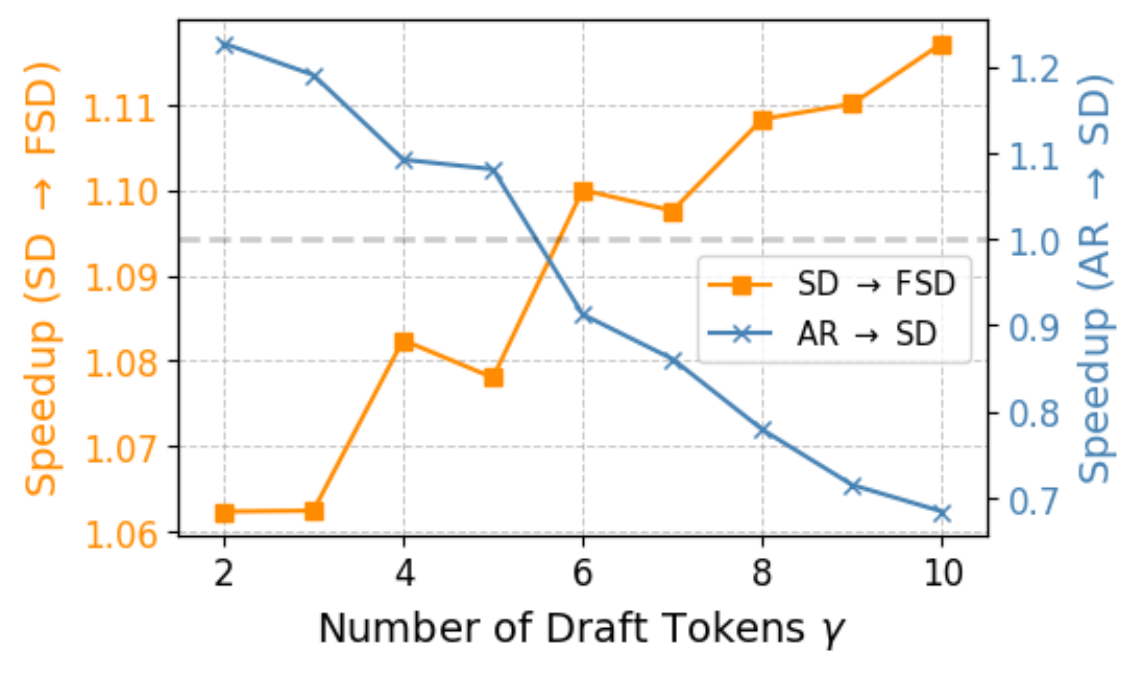}
        \caption{Effect of number of draft tokens ($\gamma$).}
        \label{fig:effect_of_gamma}
    \end{subfigure}
    \caption{Ablation studies: (a) Effect of varying the number of early exit threads, (b) Effect of varying the number of draft tokens ($\gamma$).}
    \label{fig:ablation_study}
    \vspace{-2mm}
\end{figure}

\subsection{Ablation Studies}
\textbf{Effect of Number of Threads:}
In Figure \ref{fig:effect_threads}, we show the speedup of our FSD relative to vanilla SD and the cache miss rate as the number of early exit threads increases up to 30. Using the GSM8K dataset with the Vicuna-68M/Vicuna-7B models on an RTX client, we find that the cache miss rate decreases as more threads are added, improving speedup. However, after around 15 threads, the speedup begins to plateau, and further increases in thread count yield minimal additional speedup. This is because the priority queues process the most promising early exits first, making it more likely to match the final output with the initial threads rather than the later ones.

% \begin{figure}[t]
%     \centering
%     \includegraphics[width=0.7\linewidth]{figures/gamma_effect_v1.pdf}
%     \caption{Effect of number of draft tokens ($\gamma$).}
%     \label{fig:effect_of_gamma}
%     %\vspace{-3mm}
    
% \end{figure}

\noindent
\textbf{Effect of $\gamma$:}
The number of tokens, $\gamma$, significantly influences the efficiency of speculative decoding. In Fig. \ref{fig:effect_of_gamma}, we plot speedup between SD and FSD, and between autoregressive (AR) and SD, as $\gamma$ increases up to 10. We use the GSM8K dataset with the Vicuna-68M/Vicuna-7B models on an RTX-based client. Our FSD method shows greater latency improvements compared to vanilla SD as $\gamma$ increases, enhancing the benefits of pre-drafting. However, an excessively large $\gamma$ can hinder the speculative decoding process, causing the overall speedup to decrease. As shown, the speedup of SD relative to AR falls below 1x when $\gamma$ exceeds 5.

\begin{wraptable}{r}{0.47\textwidth}
    \centering
    \vspace{-1em}
    \caption{Ablation study of client-server queue strategies. $r$ indicates the average response time (lower is better) for different queue lengths.}
    \label{tab:q_ablation_study}
    \begin{tabular}{@{}llccc@{}}
        \toprule
        \textbf{Client} & \textbf{Server} & \textbf{$r$ (3T)} & \textbf{$r$ (5T)} & \textbf{$r$ (10T)} \\
        \midrule
        \multirow{2}{*}{Priority} & Priority & \textbf{79.85} & \textbf{62.93} & 27.57 \\
                                  & FIFO     & 80.40          & 63.48          & 27.73 \\
        \midrule
        \multirow{2}{*}{Random}   & Priority & 82.32          & 63.25          & \textbf{26.92} \\
                                  & FIFO     & 84.56          & 65.36          & 27.68 \\
        \midrule
        \multirow{2}{*}{FIFO}     & Priority & 82.61          & 64.43          & 27.69 \\
                                  & FIFO     & 85.43          & 65.67          & 27.84 \\
        \bottomrule
    \end{tabular}
    \vspace{-1em}
\end{wraptable}

\noindent
\textbf{Importance of Priority Queue:}
Since our system is asynchronous, we need queues for graceful operation. Further, we organize the queues in priority, determined by the confidence score of the generated token (Eq. \ref{eq:confidence}). This prioritization is especially beneficial when the number of threads is limited. Table \ref{tab:q_ablation_study} presents an ablation study comparing different queue configurations. 
On the server side, we use either a priority queue or a FIFO queue, while on the client side, we also include a random queue as an option. We report the cache miss rate ($r$) for systems with 3, 5, and 10 threads, denoted as 3T, 5T, and 10T, respectively.

Our findings indicate that when the server uses a priority queue, it significantly improves performance for any given queue type on the client, although this benefit decreases with a higher thread count. On the client side, a priority queue consistently outperforms both the random and FIFO queues.

% \begin{table}[t]
%     \centering
%     \caption{Ablation Study of queue choices.}
%     \label{tab:q_ablation_study}
%     \begin{tabular}{@{}llccc@{}}
%         \toprule
%         \textbf{Client} & \textbf{Server} & \textbf{$r$ (3T)} & \textbf{$r$ (5T)} & \textbf{$r$ (10T)} \\ \midrule
%         \multirow{2}{*}{Priority} & Priority   &   \textbf{79.85}      &  \textbf{62.93}        & {27.57} \\
%                                   & FIFO       &    80.40     & 63.48        & 27.73  \\ \midrule
%         \multirow{2}{*}{Random}   & Priority   &   82.32      & 63.25   & \textbf{26.92} \\
%                                   & FIFO       &     84.56    & 65.36   & 27.68 \\ \midrule
%         \multirow{2}{*}{FIFO}     & Priority   &    82.61     & 64.43   & 27.69 \\
%                                   & FIFO       &    85.43     & 65.67   & 27.84 \\ \bottomrule
%     \end{tabular}
%     % %\vspace{-3mm}
    
% \end{table}

\subsection{Robotics Case Study: Vision-Language Navigation on Unitree Go2}

% Introduce the platform
To evaluate the real-world applicability of our approach, we deploy the edge-cloud speculative decoding system on the Unitree Go2 EDU quadruped robot. This platform features an onboard NVIDIA Jetson Orin board, which includes an 8-core ARM Cortex-A78AE v8.2 64-bit CPU and 16GB of 128-bit LPDDR5 unified memory, offering up to 157 TOPS of compute.  Communication between the robot and the server is established over Wi-Fi 6.

% Introduce the method
The robot receives natural language instructions such as “go to the red chair” or “turn left at the hallway” and uses its front-facing RGB camera to perceive the environment. A vision-language model (VLM) processes the visual observations and language commands to generate mid-level navigation actions (e.g., move forward small/medium/large), following the approach of \citet{cheng2024navila}. These actions are then executed by the robot’s onboard controller. To enhance decision quality and interpretability, we additionally prompt the VLM to provide reasoning alongside its action outputs.

\begin{figure}
\vspace{-8mm}
    \centering
    \includegraphics[width=\linewidth]{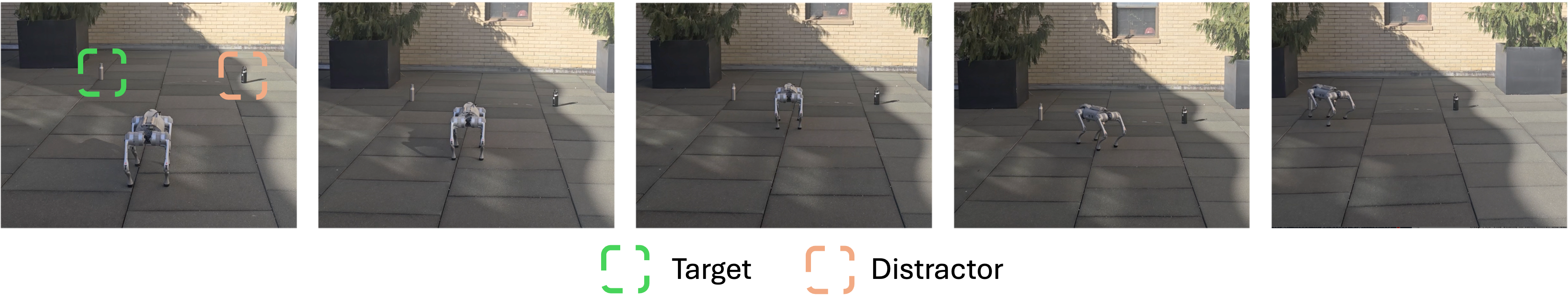}
    \caption{Example run of the Unitree Go2 robot performing an object-finding task using vision-language-based control. The robot receives the instruction \textit{“find the silver bottle”} and navigates the environment while distinguishing the correct object from a similar distractor.}
    \label{fig:bot_progress}
\end{figure}

We deploy a quantized version of Qwen-2-VL-2B as the on-device draft model and offload token verification to the full-size Qwen-2-VL-7B model hosted on an A100 GPU in the cloud. Figure~\ref{fig:bot_progress} illustrates an example scenario in which the robot is instructed to locate a specific object—in this case, a silver bottle. To test the model’s reasoning and grounding capabilities, we introduce a distractor object of similar appearance. The robot successfully navigates the environment and identifies the correct object, demonstrating the effectiveness of our method on a vision-language-based control task.

% Show results
Table~\ref{tab:system_metrics_robot}(a) reports key system-level metrics from our deployment, including drafting and verification latencies, communication overhead, and peak GPU memory usage. Table~\ref{tab:system_metrics_robot}(b) highlights the performance improvements enabled by our method, showing speedups from standard autoregressive decoding (AR) to edge-cloud speculative decoding (SD) and further to Fast Speculative Decoding (FSD). It also includes average accepted tokens per round, cache miss rate, and average early exits. Overall, our system achieves a \textbf{21\%} speedup over conventional cloud-based autoregressive decoding, validating the practicality of our approach for real-time, language-conditioned robot control on resource-constrained edge platforms. 

% \begin{table*}[t]
%   \centering
%   \caption{Average system metrics for evaluation on Unitree Go2 client and A100 Server.}
%   \label{tab:system_metrics_robot}
%   % \vskip 0.1in
%     \begin{tabular}{l c}
%     \toprule
%     \textbf{Metric} & Value \\
    
%     \midrule
%       Drafting Latency ($\gamma T_p, \gamma = 4$) & 288ms  \\
%       Verification Latency ($T_q$) & 620ms \\
%       Latency Ratio ($c = T_p/T_q$) & 0.11 \\
%       Communication Latency ($T_{c}$) & 120ms \\
%       Max GPU memory   & 12.4G \\
%       Num EE Threads   & 6 \\
%       \midrule
%       Avg Tokens ($\tau$) ($\uparrow$) & 2.92 \\
%       Cache miss rate ($\downarrow$) & 55.63\\
%       Avg EE ($\downarrow$) & 8 \\
%       Speedup AR $\rightarrow$ SD ($\uparrow$) & 0.90x \\
%       Speedup SD $\rightarrow$ FSD ($\uparrow$) & \textbf{1.34x} \\

%     \bottomrule
%     \end{tabular}%
%     % %\vspace{-3mm}
  
% \end{table*}

\begin{table*}[t]
\vspace{-2mm}
  \centering
  \caption{System-level evaluation metrics for our edge-cloud speculative decoding setup on the Unitree Go2 robot (Jetson Orin) and A100 server. (a) Reports core system and latency metrics. (b) Summarizes the performance gains with our FSD method.}
  \label{tab:system_metrics_robot}
  \begin{minipage}{0.48\textwidth}
    \centering
    \begin{tabular}{l c}
    \toprule
    \textbf{Metric} & \textbf{Value} \\
    \midrule
    Drafting Latency ($\gamma T_p$, $\gamma = 4$) & 288ms \\
    Verification Latency ($T_q$) & 620ms \\
    Latency Ratio ($c = T_p/T_q$) & 0.11 \\
    Communication Latency ($T_c$) & 120ms \\
    Max GPU Memory & 12.4G \\
    Num EE Threads & 6 \\
    \bottomrule
    \end{tabular}
    \caption*{(a) System and latency metrics}
  \end{minipage}
  \hfill
  \begin{minipage}{0.48\textwidth}
    \centering
    \begin{tabular}{l c}
    \toprule
    \textbf{Metric} & \textbf{Value} \\
    \midrule
    Avg Tokens ($\tau$) ($\uparrow$) & 2.92 \\
    Cache Miss Rate ($\downarrow$) & 55.63 \\
    Avg EE ($\downarrow$) & 8 \\
    Speedup AR $\rightarrow$ SD ($\uparrow$) & 0.90x \\
    Speedup SD $\rightarrow$ FSD ($\uparrow$) & \textbf{1.34x} \\
    \bottomrule
    \end{tabular}
    \caption*{(b) FSD speedup and performance metrics}
  \end{minipage}
  \vspace{-5mm}
\end{table*}

\section{Related Work}
% On-device/ Edge-cloud distributed LLMs.

There is a significant interest in enabling edge devices to run LLMs. 
The deployment of collaborative AI inference systems across the edge and the cloud introduces unique challenges such as latency constraints, bandwidth limitations, and inconsistent network conditions. 
One straightforward approach is to design smaller models \cite{lu2024small}. While all the popular class of models such as OPT \cite{zhang2022opt}, Llama \cite{touvron2023llama, dubey2024llama}, and Gemma \cite{team2024gemma} have smaller scale models, they are either not small enough to run on an edge device or not accurate enough to reliably deploy in practical applications.

Quantization is one of the heavily focused methods to enable on-device LLMs \cite{lin2024awq}. \citet{yu2024edge} aim to compress the models with layer-wise pruning and quantization to enable edge LLMs. \citet{qu2024mobile} discuss an approach of enabling LLMs to run on 6g edge devices. On system side, \cite{xu2024empowering} focus on leveraging on-device Neural Processing Unit (NPU).

Early exit strategies, which allow intermediate layers of deep networks to make predictions without waiting for the full forward pass, have been extensively explored for resource-constrained devices. Pioneering works such as Conditional Deep Learning \cite{panda2016conditional} and BranchyNet \cite{teerapittayanon2016branchynet} introduced the idea of adding multiple exit points to deep neural networks to reduce computation time.
Recent research has also explored layer skipping in LLMs for enhanced efficiency \cite{fan2024not}, with dynamic compute allocation based on tokens \cite{raposo2024mixture}.
In terms of multi-device speculative decoding, \cite{mcdanel2024amusd} has recently shown that asynchronous speculative decoding over multiple GPUs can be beneficial. However, it uses shared memory to communicate between devices, so it is not directly applicable to edge-cloud scenarios. 

Our work lies at the intersection of LLM decoding, early exit mechanisms, and distributed inference optimization, addressing a critical gap by proposing a preemptive speculative decoding framework tailored for edge-cloud environments.
To the best of our knowledge, this is the first work to show end-to-end speculative decoding with models split between edge and cloud. Further, we comprehensively analyze and demonstrate the system-level trade-offs during the implementation of collaborative edge-cloud decoding, which no prior work has investigated.

\section{Conclusion and Discussion}
We introduced a novel speculative edge-cloud decoding framework, offering a cost-effective alternative to traditional cloud-based deployment. By distributing the draft and target models between edge and server environments, our solution significantly reduces high API costs. Early exits and pre-drafting allow us to enhance parallelism by leveraging idle client time and reducing server idle time. Our comprehensive end-to-end evaluation on the NVIDIA Jetson Nano highlights the feasibility of efficient edge-cloud collaborative LLM inference on resource-limited edge devices. On Jetson Nano, speculative edge-cloud decoding achieves up to a 35\% speedup over cloud-based autoregressive decoding, with up to an additional 11\% performance gain enabled by pre-drafting and early exits. Further, we validate our approach with execution of vision language models on the Unitree Go2 quadruped robot. We achieve an overall 21\% speedup over standard cloud-based autoregressive decoding, demonstrating the effectiveness and real-world applicability of our framework for robotics use cases.

% We propose a method for performing server-client collaborative speculative decoding. We demonstrate the effectiveness with Jetson Nano device. 

Our method operates effectively without making assumptions about communication delays, and we show that it remains practical under real-world conditions. While communication latency can be a limiting factor in extreme cases, our use of priority queues helps optimize bandwidth usage. For scenarios with constrained network conditions, further tuning and adaptive scheduling policies offer promising avenues to enhance performance. Similarly, while pre-drafting leverages idle client time for parallelism, it introduces modest compute overhead on the client. This is manageable for most edge platforms, and the ability to scale the number of threads provides a flexible trade-off between latency and energy efficiency, especially for battery-powered devices.

Our proof-of-concept is implemented in Python, which, while convenient for experimentation, leaves room for further optimization. A low-level C++ implementation with shared memory could substantially improve performance, making the system even more suitable for latency-sensitive applications. Currently, our system supports single-client interaction with the server, but extending it to support multi-client concurrency is a natural next step. We envision future work enabling scalable, concurrent edge-cloud inference with early exits, making our approach even more applicable to real-world deployment scenarios.

\bibliography{main}
\bibliographystyle{tmlr}

\appendix
\section{System Design}\label{section:appendix_system_design}
\textbf{Client:}
On the client side, the primary goal is to maximize idle time usage and increase cache hit rates. As shown in Algorithm \ref{alg:clientalgo}, the client maintains a priority queue \(Q_p\), a pre-draft cache \(C\), and a \textsc{Receiver} thread. After sending draft tokens to the server for verification, the \textsc{Receiver} thread listens for server callbacks, which provide outputs from early exits.

Since the client’s bottleneck lies in the processing power required for generating pre-draft tokens, it is essential to prioritize the handling of early exit outputs. The priority queue $Q_p$ organizes these outputs according to their confidence levels (Eq. \ref{eq:confidence}), prioritizing the most promising ones for pre-drafting. It is populated asynchronously as early exit outputs are received by the client. If the client's device has multiple available threads, it can process several early exits in parallel to generate more pre-draft tokens. All the pre-draft tokens are stored in the pre-draft cache. Once the client receives the final exit output, it checks the pre-draft cache for the corresponding tokens. If there is a cache hit, the pre-drafted tokens are sent to the server immediately for the next verification round.

% Explain confidence as priority

\noindent
\textbf{Server:}
As detailed in Algorithm \ref{alg:serveralgo}.
the server consists of two asynchronous threads: \textsc{Listener} and \textsc{Sender}. The \textsc{Listener} processes the verification requests from the client. It takes in the prefix \( x_{1:t} \), draft tokens \( x_{t: t+ \gamma} \), and their corresponding probability distribution \( p_{1:\gamma} \).

As shown in Fig. \ref{fig:main_fig}, communication typically becomes the bottleneck in the server as early exit outputs are produced faster than the network can transmit. Early exit outputs are placed in a queue on the server side and transmitted sequentially to handle this. Let \( {Q}_e \) represent the server-side queue storing the early exit outputs:
\begin{equation}  
    {Q}_{e} = \{ x_{t:t + \delta^{(1)} + 1}^{(1)}, \ldots, x_{t:t + \delta^{(L)} + 1}^{(L)} \}.  
\end{equation}  
Asynchronously, the \textsc{Sender} thread sends the early exit outputs from the queue based on priority determined by the confidence score (Eq. \ref{eq:confidence}).

\begin{algorithm}[h!]
\caption{Client-Side Algorithm}
\label{alg:clientalgo}
\begin{algorithmic}[1]
\State \textbf{Initialize:} Draft model \(\mathcal{M}_{p}\), Queue \(Q_p\), Cache \(C\), \textsc{Receiver}
\State \textbf{Input:} Prefix \(x_{1:t}\), \# Total Tokens \(T\), \# Draft tokens \(\gamma\)
\State \textbf{Output:} Final tokens \(x_{t+1:T}\)

\For{$i = 1$ to \(\gamma\)} 
    \State $x_{t + i}, p_{i} \gets \textsc{Draft}(\mathcal{M}_{p}, x_{1:t + i - 1})$
\EndFor
\State \textsc{Send}$(x_{1:t + \gamma}, p_{1:\gamma})$

\While{$t < T$}
    \While{$Q_p$ not empty}
        \State $x'_{t+1:t+\delta'+1}, s' \gets Q_p.\text{pop()}$
        \If{$x'_{t+1:t+\delta'+1}$ not in $C$}
            \State $C[x'_{t+1:t+\delta'+1}] \gets \textsc{PreDraft}(x_{1:t}, x'_{t+1:t+\delta'+1})$
        \EndIf
    \EndWhile

    \State $y_{1:t'} \gets \text{Concat}(x_{1:t}, x_{t+1:t+\delta+1})$
    \If{$x_{t+1:t+\delta+1} \in C$}
        \State $y_{t':t' + \gamma}, p_{1:\gamma} \gets C[x_{t+1:t+\delta+1}]$
    \Else
        \For{$i = 1$ to \(\gamma\)}
            \State $y_{t' + i}, p_{i} \gets \textsc{Draft}(\mathcal{M}_{p}, y_{1:t' + i - 1})$
        \EndFor
    \EndIf

    \State \textsc{Send}$(y_{t':t' + \gamma}, p_{1:\gamma})$
    \State $t \gets t'$, $x \gets y$, $C.\text{reset}()$, $Q_p.\text{reset}()$
\EndWhile
\Statex
\Function{PreDraft}{} \funclabel{alg:pre-draft}
    \State \textbf{Input:} Prefix $x_{1:t}$, Tokens $x'_{t+1:t+\delta'+1}$
    \State $y_{1:t'} \gets \text{Concat}(x_{1:t}, x'_{t+1:t+\delta'+1})$
    \For{$i = 1$ to \(\gamma\)} 
        \State $y_{t' + i}, p'_{i} \gets \textsc{Draft}(\mathcal{M}_{p}, x'_{1:t' + i - 1})$
    \EndFor
    \State \Return $y_{t':t' + \gamma}, p'_{1:\gamma}$
\EndFunction

\Statex
\Function{Receiver}{} \funclabel{alg:receiver}
    \State \textbf{Input:} Tokens $x'_{t+1:t+\delta'+1}$, Priority $s'$, \textit{isfinalexit}
    \If{\textit{isfinalexit}}
        \State $x_{t+1:t+\delta+1} \gets x'_{t+1:t+\delta'+1}$
    \Else
        \State $Q_p.\text{push}(x'_{t+1:t+\delta'+1}, s')$
    \EndIf
\EndFunction
\end{algorithmic}
\end{algorithm}

\begin{algorithm}[h!]
\caption{Server-Side Algorithm}
\label{alg:serveralgo}
\begin{algorithmic}[1]
\State \textbf{Initialize:} Target Model \(\mathcal{M}_{q}\), Verification criterion \textsc{Verify}, Queue $Q_e$, \textsc{Listener}, \textsc{Sender}

\Function{Listener}{} \funclabel{alg:listener}
    \State \textbf{Input:} Prefix and draft tokens $x_{1:t+\gamma}$, Probs. $p_{1:\gamma}$
    \State $x^{(1:L)}_{t+1:t+\delta+1}, q^{(1:L)}_{1:\delta+1} \gets \textsc{Verify}(\mathcal{M}_{q}, x_{1:t+\gamma}, p_{1:\gamma})$
    \ForAll{$i = 1, \dots, L - 1$} 
        \State $s^{(i)} \gets \text{max}(q^{(i)}_{1:\delta+1})$
        \State $Q_e.\text{push}(x^{(i)}_{t+1:t+\delta+1}, s^{(i)})$
    \EndFor
    \State \textsc{Send}$(x^{(L)}_{t+1:t+\delta+1}, \text{isfinalexit} = \text{True})$
    \State $Q_e.\text{reset}()$
\EndFunction

\Function{Sender}{} \funclabel{alg:sender}
    \While{$Q_e$ not empty}
        \State $(x'_{t+1:t+\delta+1}, s') \gets Q_e.\text{pop()}$
        \State \textsc{Send}$(x'_{t+1:t+\delta+1}, s', \text{isfinalexit} = \text{False})$
    \EndWhile
\EndFunction

\end{algorithmic}
\end{algorithm}

\section{Speedup Projection Analysis}\label{section:appendix_comm_projection}

\begin{figure*}[h!]
    \centering
    \includegraphics[width=0.85\linewidth]{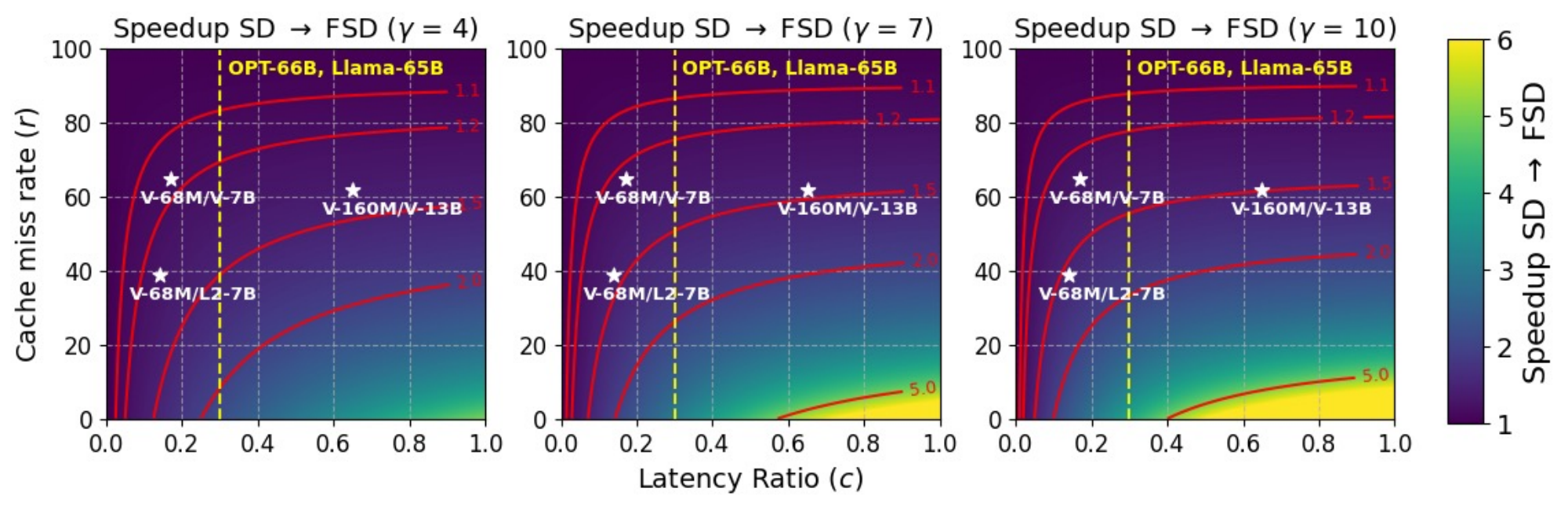}
    \caption{Estimated speedup per round of our FSD relative to vanilla SD method assuming no communication latency. We emphasize the contours for speedups of 1.1x, 1.2x, 1.5x, 2x, and 5x, and we indicate the position of our models within this landscape. Additionally, we highlight the operating range for the larger models OPT-66B and Llama-65B.}
    \label{fig:analysis}
    %\vspace{-3mm}
\end{figure*}

One of the bottlenecks in our system is communication between the devices. In addition to directly adding to the latency, it also bottlenecks the number of early exit verifications communicated back to the client. This increases the cache miss rate further increasing the latency.
While this is a challenge at present due to limited communication network capabilities, several works have shown the vision of having edge LLMs on 6g networks with a projected network speed up to 10 Tbps \cite{banafaa20236g, lin2023pushing, xu2024large, friha2024llm, qu2024mobile, zhang2024edgeshard}.
In scenarios where communication latencies \(T_{c}\) is negligible relative to drafting and verification time, and ignoring thread synchronization latency ($T_r$) for simplicity, we can approximate the speedup SD $\rightarrow$ FSD as:
\begin{equation}
    \text{Speedup (SD} \rightarrow \text{FSD)} = \frac{\gamma c + 1}{r(\gamma c) + 1}
    \label{eq:speedup_proj}
\end{equation}

This formula reduces the final speedup to be affected by two factors---Cache miss rate \(r\) and Latency Ratio \(c\). 
%### Analysis of Speedup Trends and Operating Range
Cache miss rate $r$ depends on the redundancy in the target model and how well the early exit adapters are trained.
% When \(r \rightarrow 0\) (high cache hit rate), our FSD method achieves near-maximum speedup, as most predictions from the draft model are verified at an early exit and the pre-drafted tokens are readily available as soon as the final exit output is available to the client. 
% In standard speculative decoding typically \(c \ll 1\) because draft model is considered to be much faster than the target model. However, in the case of edge devices, since the client devices have much smaller processing power, it pushes \(c\) to a higher value.
On the other hand, $c$ is highly dependent on the compute capability of the edge device.
Since edge devices are often slower, this pushes the $c$ to be higher. 

We visualize Eq. \ref{eq:speedup_proj} as a heatmap in Fig. \ref{fig:analysis} for $\gamma$ values 4, 7 and 10. For reference, we plot the measured $c$ and $r$ values based on the Jetson implementation of our current set of models within this landscape.
Naturally, having a lower $r$ will improve speedup, but the usefulness of our FSD method becomes more pronounced as we get to higher $c$ and $\gamma$.
For model sets with a latency ratio greater than 0.5 and well-trained early exit adapters that achieve a cache miss rate of less than 10\%, we can anticipate a speedup over 5x.
% For example, the Vicuna-160M/Vicuna-13B model has a  in Fig. \ref{fig:analysis}.

% In Fig. \ref{fig:analysis}, we provide empirical estimates of the speedup for different values of \(c\) and \(r\) for a set of \(\gamma\) values, illustrating the practical operating range. 
% With low communication latency, the upper bound on speedup is mainly governed by the ratio \(c\) and the ability to keep the cache miss rate \(r\) minimal. We highlight the contours of 1.1x, 1.2x, 1.5x, 2x, and 5x speedup.
% We also mark the projected speedup of the current set of models given their latency ratio $c$ and the observed cache miss rate ($r$) based on Jetson to see what is the potential speedup they can achieve. For example, with Vicuna-68M/Llama2-7B model pair, for $\gamma$ = 4, while currently we report 1.11x speedup, given faster communication, we expect a speedup of more than 1.2x. Further, with $\gamma =7, 10$, the expected speedup can increase up to 1.5x-2x.

Extending our analysis to larger models, specifically OPT-66B and Llama-65B with draft models OPT-125M and NoFT-Wide-796M, we use reported latencies from \cite{yan2024decoding} (6.6 ms for draft, 67 ms for target) and factor in a 3x slowdown on Jetson, arriving at $c \approx 0.3$. This value is illustrated by the yellow line in Fig. \ref{fig:analysis}. For instance, to achieve a 2x speedup with $\gamma$ values of 4, 7, and 10, the cache miss rate must remain below 10\%, 25\%, and 35\%, respectively.

\section{Batch Processing}
Table \ref{tab:latency_comparison} shows latency analysis for Vicuna-7B (A100) and Vicuna-68M (Jetson Nano). Batch processing improves throughput but not always latency; e.g., batch size 32 increases A100 latency by over 5x. However, API providers often offer discounts for batch processing (e.g., OpenAI provides 50\% discount \href{https://openai.com/api/pricing/}{OpenAI Pricing}), making it a cost-saving approach. On the client, batch processing shows a smaller latency increase—batch sizes of 1 and 8 differ by 15\%. A batched pre-drafting approach could reduce latency but requires waiting to accumulate multiple early exits, introducing a trade-off.

\begin{table}[h!]
\centering
\caption{Server and client latency for different batch sizes.}
\begin{tabular}{ccc}
\hline
\textbf{Batch} & \textbf{Server Latency} & \textbf{Client Latency} \\ 
\textbf{Size}  & \textbf{(Vicuna-7B)}    & \textbf{(Vicuna-68M)}  \\ 
               & \textbf{(A100)}         & \textbf{(Jetson)}      \\ \hline
1  & 17.94  & 7.76  \\ 
2  & 21.75  & 8.02  \\ 
3  & 24.35  & 7.57  \\ 
4  & 24.63  & 7.87  \\ 
8  & 30.73  & 8.97  \\ 
12 & 44.52  & 9.71  \\ 
16 & 45.79  & 10.63 \\ 
32 & 96.94  & 14.49 \\ 
64 & 258.79 & OOM   \\ \hline
\end{tabular}
\label{tab:latency_comparison}
\end{table}

% \section{Robotics Case Study}

\end{document}

%% file: math_commands.tex
\usepackage{amsmath,amsfonts,bm}

\def\eqref#1{equation~\ref{#1}}
\def\1{\bm{1}}

\def\ra{{\textnormal{a}}}

\def\rx{{\textnormal{x}}}

\def\rva{{\mathbf{a}}}

\def\erva{{\textnormal{a}}}

\def\ervx{{\textnormal{x}}}

\def\rmA{{\mathbf{A}}}

\def\vmu{{\bm{\mu}}}
\def\vtheta{{\bm{\theta}}}
\def\va{{\bm{a}}}

\def\ve{{\bm{e}}}

\def\vx{{\bm{x}}}

\def\eva{{a}}

\def\mA{{\bm{A}}}

\def\mH{{\bm{H}}}
\def\mI{{\bm{I}}}
\def\mJ{{\bm{J}}}

\def\mX{{\bm{X}}}

\def\mSigma{{\bm{\Sigma}}}

\DeclareMathAlphabet{\mathsfit}{\encodingdefault}{\sfdefault}{m}{sl}
\SetMathAlphabet{\mathsfit}{bold}{\encodingdefault}{\sfdefault}{bx}{n}
\newcommand{\tens}[1]{\bm{\mathsfit{#1}}}
\def\tA{{\tens{A}}}

\def\tX{{\tens{X}}}

\def\gG{{\mathcal{G}}}

\def\sA{{\mathbb{A}}}
\def\sB{{\mathbb{B}}}

\def\sS{{\mathbb{S}}}

\def\emA{{A}}

\newcommand{\etens}[1]{\mathsfit{#1}}

\def\etA{{\etens{A}}}

\newcommand{\E}{\mathbb{E}}

\newcommand{\R}{\mathbb{R}}

\newcommand{\KL}{D_{\mathrm{KL}}}
\newcommand{\Var}{\mathrm{Var}}

\newcommand{\Cov}{\mathrm{Cov}}
\newcommand{\normltwo}{L^2}
\newcommand{\normlp}{L^p}

\newcommand{\parents}{Pa} % See usage in notation.tex. Chosen to match Daphne's book.